\documentclass{article}

\usepackage[table,xcdraw]{xcolor} 
\usepackage{arxiv}
\usepackage[utf8]{inputenc} 
\usepackage[T1]{fontenc}    
\usepackage{hyperref}       
\usepackage{url}            
\usepackage{booktabs}       
\usepackage{amsfonts}       
\usepackage{nicefrac}       
\usepackage{microtype}      
\usepackage{graphicx}
\usepackage{natbib}
\usepackage{doi}
\usepackage{enumitem} 
\usepackage{caption} 

\title{Embracing Dialectic Intersubjectivity: Coordination of Different Perspectives in Content Analysis with LLM Persona Simulation}

\author{
    \begin{tabular}{ccc}
        \begin{minipage}[t]{0.32\textwidth}
            \centering
            Taewoo Kang \\
            Department of Media and Information \\
            Michigan State University \\
            \texttt{kangtaew@msu.edu}
        \end{minipage} &
        \begin{minipage}[t]{0.32\textwidth}
            \centering
            Kjerstin Thorson \\
            College of Liberal Arts \\
            Colorado State University \\
            \texttt{k.thorson@colostate.edu}
        \end{minipage} &
        \begin{minipage}[t]{0.32\textwidth}
            \centering
            Tai-Quan Peng \\
            Department of Communication \\
            Michigan State University \\
           \texttt{pengtaiq@msu.edu}
        \end{minipage} \\
        \\[1em] 
        \begin{minipage}[t]{0.32\textwidth}
            \centering
            Dan Hiaeshutter-Rice \\
            Department of Advertising and Public Relations \\
            Michigan State University \\
            \texttt{dhrice@msu.edu}
        \end{minipage} &
        \begin{minipage}[t]{0.32\textwidth}
            \centering
            Sanguk Lee \\
            Department of Communication Studies \\
            Texas Christian University \\
            \texttt{sanguk.lee@tcu.edu}
        \end{minipage} &
        \begin{minipage}[t]{0.32\textwidth}
            \centering
            Stuart Soroka \\
            Departments of Communication and Political Science \\
            University of California, Los Angeles \\
            \texttt{snsoroka@ucla.edu}
        \end{minipage}
    \end{tabular}
}

\date{\today}

\renewcommand{\shorttitle}{Kang et al. (2025) Embracing dialectic intersubjectivity}

\hypersetup{
    pdftitle={Embracing Dialectic Intersubjectivity},
    pdfsubject={q-bio.NC, q-bio.QM},
    pdfauthor={Taiwoo Kang, Kjerstin Thorson, Tai-Quan Peng, Dan Hiaeshutter-Rice, Sanguk Lee, Stuart Soroka},
    pdfkeywords={large language models, text analysis, agent-based modeling, intersubjectivity, political polarization}
}

\begin{document}
\maketitle
\begin{abstract}
	This study attempts to advancing content analysis methodology from consensus-oriented to coordination-oriented practices, thereby embracing diverse coding outputs and exploring the dynamics among differential perspectives. As an exploratory investigation of this approach, we evaluate six GPT-4o configurations to analyze sentiment in Fox News and MSNBC transcripts on Biden and Trump during the 2020 U.S. presidential campaign, examining patterns across these models. By assessing each model’s alignment with ideological perspectives, we explore how partisan selective processing could be identified in LLM-Assisted Content Analysis (LACA). Findings reveal that partisan persona LLMs exhibit stronger ideological biases when processing politically congruent content. Additionally, intercoder reliability is higher among same-partisan personas compared to cross-partisan pairs. This approach enhances the nuanced understanding of LLM outputs and advances the integrity of AI-driven social science research, enabling simulations of real-world implications.
\end{abstract}


\section{Introduction}
Large Language Models (LLMs), such as ChatGPT, allow researchers to analyze extensive text corpora with efficiency \citep{Bail2024}. However, human and algorithmic biases can influence their outputs, reflecting ideological or cultural skewness in the training data \citep{Bender2021, Kroon2023}. With the growing adoption of LLM-Assisted Content Analysis (LACA)—a method replacing manual coding with LLM-generated datasets \citep{Chew2023}—the scientific community must ensure these tools enhance understanding rather than reinforce existing biases \citep{Messeri2024}. \par

Human bias is a long-standing issue in content analysis, which traditionally depends on a consensus-oriented research practice. Intercoder agreement is used to ensure reliability, aiming for shared interpretations of text that can be statistically validated \citep{Krippendorff1999}. However, this emphasis on consensus may inadvertently neglect diverse perspectives, favoring uniform interpretations that risk simplifying the sociocultural complexities embedded in textual connotations. This is particularly true in studies focused on ideologically polarized content, as social groups with differing ideologies interpret messages distinctly based on their belief systems \citep{neuendorf2015, Schiffer2006}. Equating reliability with coder agreement can therefore overlook alternative valid interpretations, leading to a form of reliable invalidity—where consistent measurements obscure the true diversity of a text’s meanings. \par

Given the interplay of human and algorithmic biases in LACA, we propose adopting dialectic intersubjectivity \citep{Matusov1996}. This framework reframes consensus as one of many possible outcomes, emphasizing the coordination of diverse perspectives \citep{Gillespie2010}. By valuing both agreement and productive disagreement, it positions content analysis as a means of understanding the relational dynamics among differing standpoints. This approach is especially relevant for social scientific uses of LLMs, which can generate diverse outputs by simulating various sociocultural or ideological personas \citep{Argyle2023, Santurkar2024}. Applying dialectic intersubjectivity to LACA could enable researchers to examine the range of interpretations generated by LLMs, offering insights into how these tools reflect the complexities of real-world perspectives. \par

We advance this approach by employing six LLM-simulated partisan personas to analyze sentiment toward the 2020 U.S. presidential candidates, Joe Biden and Donald Trump, in Fox News and MSNBC. By conducting comparative analyses across the personas, we aim to assess the degree to which each model reflects partisan biases, with particular attention to the extent to which an LLM mirrors typical standpoints of different ideological groups. Each model’s output was evaluated in terms of positivity and negativity toward both candidates, with intercoder reliability testing both within-community (same partisan personas) and between-community (cross-partisan personas) model pairs. \par

The findings reveal that while LLMs can effectively simulate partisan personas—showing higher coder agreement for within-community than for between-community pairs, they exhibit notable sentiment divergence when analyzing ideologically like-minded content (i.e., Democrat personas analyzing MSNBC and Republican personas analyzing Fox News). This tendency contrasts with more convergence when analyzing ideologically cross-cutting content, highlighting the critical implications of dialectic intersubjectivity in LACA for political text; that is, moving beyond consensus-oriented metrics allows for a holistic capture of the diverse views. This shift promotes a more nuanced approach to LACA that acknowledges varied interpretations, allowing LLMs to enhance rather than constrain the plurality of human standpoints.

\section{Establishing Reliability in Content Analysis: Pros and Cons}
\label{sec:headings }

Content analysis is “the backbone of communication research” \citep{Krippendorff1999}. Identifying the properties of messages is crucial for examining their effects. As a methodology for capturing the message states, replicability is the scientific underpinning of content analysis, with the reliability of measurement determining the overall quality of the research \citep{Kolbe1991}. However, while reliability is necessary, it is not sufficient \citep{neuendorf2011}. \cite{Krippendorff1999} clarifies in his textbook of content analysis that while reliability ensures a study is replicable, it does not necessarily ensure that the study accurately reflects reality: “[V]alidity concerns truths,” whereas “reliability does not guarantee validity” (pp. 212-213, italic in original). This common issue in any scientific research is particularly relevant in content analysis when determining which aspects of a message are being measured. \par

Historically, systematic content analysis, developed after World War II, emerged from a need to understand how messages from key press or political elites influence the masses \citep{Berelson1952}. In this context, content analysis has evolved to primarily seek to grasp a common ground of a text, which, in turn, “privileges content analysts’ accounts over the readings by audience members” \cite[p. 20]{Krippendorff1999}. This approach is consistent with the traditional definition of content analysis, which involves drawing (statistical) “inferences about the states or properties of the sources of the analyzed texts” \cite[p. 35]{Osgood1959}. \par

This definition, focused on quantifying commonalities, has faced criticism for excluding the diverse perspectives integral to communication processes \citep{Shapiro1997, Holsti1969}). \cite{Krippendorff1999} acknowledges this critique, noting Berelson’s emphasis on quantification but stressing that “[r]eading is fundamentally a qualitative process” (pp. 19-20). He further argues that “if content analysts were not allowed to read texts in ways that differ from other readers, content analysis would be pointless” (p. 23). This underscores the idea that texts often contain multiple dimensions, making it unlikely for a single reliable measure to capture their full complexity. Even with reliability and validity, such measures often reflect only one aspect of a text, potentially overlooking other latent dimensions. \par

For instance, content analysis research shows how media portray politicians differently depending on the ideological orientation of the outlet. Conservative media tend to present conservative candidates in a positive light and liberal candidates negatively, whereas liberal media typically do the reverse \citep{Benoit2014, dalessio2000, neuendorf2004, Schiffer2006}. While a content analysis of such media biases offers a reliable inference about how news producers encode specific properties (i.e., news framing), it does not necessarily provide an equitable understanding of how audiences decode these properties. Conservative or liberal voters may interpret political framing within conservative or liberal outlets differently than what content analysts infer. This forms the crux of \cite{Krippendorff1999}’s critique of Berelson’s emphasis on capturing shared meanings: The properties agreed upon by content analysts who share common perspectives may represent only a partial reality or truth of a text (p. 20). The consensus of content analysts can be reliably invalid. \par

\section{Persistent Indirectness of Validity in Content Analysis}
\label{sec:headings}
The multifaceted nature of communication processes prevents \textit{“formal assessment of the validity of content analysis measures”}\citep{neuendorf2011}. It is unlikely that a content analyst can directly validate how communicators have encoded or decoded specific properties within a text \citep{Holsti1969}. Any validity measures thus are indirect and often tend to converge on face validity (\citealp{Janis1965}; see \citealp{Krippendorff1980} for a review). \par

Scholars are aware of these challenges, yet it is uncommon to find methodological solutions aimed at addressing them \citep{Matthes2008}. Instead, it is often recommended that the findings of content analysis be regarded as preliminary—either serving as a groundwork for further research or complementing other methodologies. As such, \cite{Krippendorff1999} advocates for an ethnographic approach, which connects content analysis with direct research on communicators (see also \citealp{Altheide1987}). Similarly, \cite{neuendorf2004} distinguishes content analysis from discourse analysis, which emphasizes reflexivity, cognitive processes, and critical reflection, noting that the former focuses on production, outputs, and broad generalizations—an approach described as industrial. This distinction often serves as a rationale for the necessity of subsequent discourse analysis, as content analysis tends to stimulate qualitative inquiry. \par

Consequently, challenges in establishing validity often led to an emphasis on directly quantifiable reliability measures as the primary criterion for evaluating content analysis. However, as noted, the reliability of a measure for capturing one facet does not guarantee its applicability to other aspects. This is evident in the recommendation that researchers validate content analysis within relatively homogeneous populations, thereby acknowledging the constrained empirical domain of common ground \citep{Krippendorff1980, neuendorf2011}. \par

\section{Intersubjectivity as Agreement}
\label{sec:headings}
To capture the heterogeneity of a text’s connotation, each latent dimension requires the identification of its own homogeneous common ground. Although this conclusion is logical, the field of content analysis has struggled to implement it due to its sole conceptualization of intersubjectivity. In content analysis research, intersubjectivity generally refers to the shared understanding throughout the research process \citep{neuendorf2004}. It is positioned as a substitute for the unattainable ideal of absolute objectivity, aiming to establish a consensus among researchers \citep{babbie2015}. As a result, intercoder agreement is suggested as the mutual understanding of text properties that coders or raters commonly identify, even while relying on their own mental schemas \citep{Potter1999}. \par

Taken for granted, there is nothing wrong with conceptualizing intersubjectivity as “sharing” or “having in common” \cite[p. 26]{Matusov1996}. However, this conceptualization restricts content analysis to being a type of consensus-oriented activity, treating disagreements as nuisances that should be eliminated. If separate analyses do not show increased overlap, it is seen as evidence of a failure in intersubjectivity \cite{Goncu1993}, leading the research to focus on the process by which coders’ subjectivities are unified \cite{Smolka1995}. Yet, the benefits of unification come with hidden costs: excluding not shared perspectives. \par

\textit{The meanings invoked by texts need not be shared. Although intersubjective agreement as to what an author meant to say or what a given text means would simplify a content analysis tremendously, such consensus rarely exists in fact. Demanding that analysts find a “common ground” would restrict the empirical domain of content analysis to the most trivial or “manifest aspects of communications,” on which Berelson’s definition relies, or it would restrict the use of content analysis to a small community of message producers, recipients, and analysts who happen to see the world from the same perspective.} \cite[p. 23]{Krippendorff1999}

In this regard, we propose that the evolution of content analysis beyond Berelson’s paradigm requires a confrontation with the validity challenge—particularly around notions of “common ground” that, while recognized, remain unresolved. By reconceptualizing the inertia surrounding intersubjectivity, may we establish a basis for the advancement of content analysis. \par

\section{Intersubjectivity with and without Agreement}
\label{sec:headings}
The traditional concept of intersubjectivity, defined as a state of shared individual understandings \citep{Rommetveit1979}, has long faced criticism; Instead, \cite{Matusov1996}, in a pivotal study, proposes re-envisioning intersubjectivity as a process grounded more in joint-activity than in the pursuit of interpersonal consensus. This participatory perspective defines intersubjectivity as “a process of a coordination of participants’ contributions in joint activity” (p. 25). By this definition, disagreement—manifesting as conflict or dispute—is as fundamental to intersubjectivity as agreement or resolution. Unlike traditional approaches, participatory approach argues that virtuous intersubjectivity emerges through the recognition, exploration, and, ideally, coexistence of diverse standpoints; Even in the absence of consensus, divergent perspectives lay a foundation for future coordination. Thus, intersubjectivity functions dialectically, holding different perspectives in tension, and serves as the interactive force of sociocultural activity. \par

Further articulating this approach, \cite{Matusov1996} references semiotician \cite{Lotman1988}, who emphasizes that while unifying individual standpoints is a text’s first function, it only achieves coherence by generating new meanings through the juxtaposition of distinct perspectives. In other words, intersubjective communication requires these varied perspectives to be acknowledged in advance. Hence, within this approach, the unit of analysis extends beyond individual perspectives to include the dialectical relationship between them. \par

\begin{figure}[ht]
    \centering
    \includegraphics[width=0.9\linewidth]{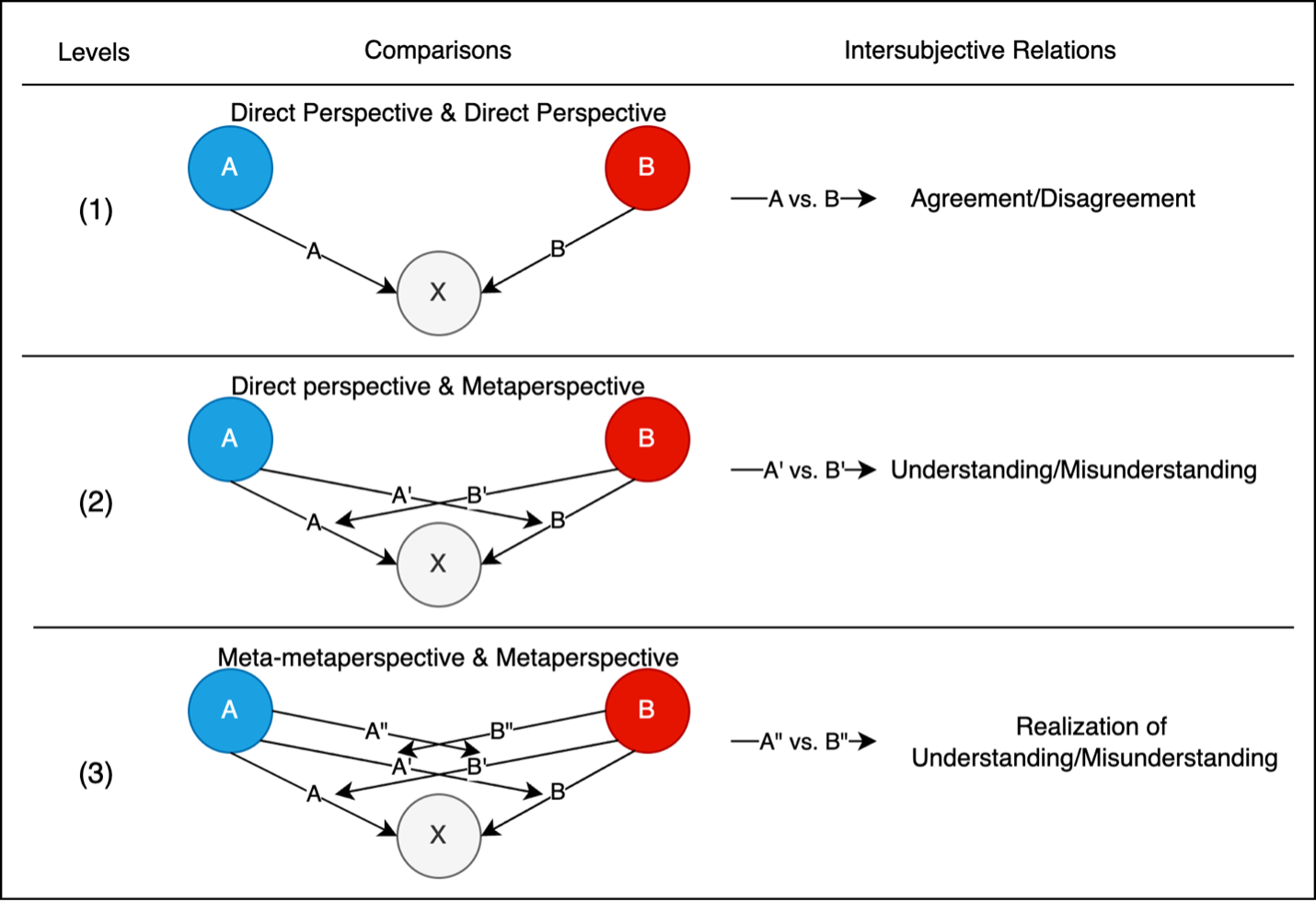}
    \caption{Three-level intersubjective comparison and relations}
    \label{fig:fig1}
    \vspace{1em} 
    \footnotesize 
    \textit{Note.} A and B: Individual content analysts; X: A given text. Created based on Gillespie and Cornish (2010, Table 2, p. 24).
\end{figure}

This dialectical relationship is elaborated in \cite{Gillespie2010}’s dialogical analysis of intersubjectivity. By defining intersubjectivity as “the variety of possible relations between people’s perspectives,” (p. 19) they capture how individuals attempt to orient themselves in response to others’ orientations \citep{Latsis2006}. Specifically, drawing on the framework by \cite{Laing1966}, they conceptualize intersubjectivity in three distinct levels, outlining possible intersubjective relations between individuals and groups (Figure~\ref{fig:fig1}). \par

The first level, the direct perspective, enables recognition of multiple self-perspectives toward a given phenomenon, allowing researchers to compare agreement and disagreement among individual views. The second level, the metaperspective, involves reflections on one’s own and others’ perspectives. At this stage, researchers can identify understanding and misunderstanding relationally across individuals. The third level, the meta-metaperspective, captures one’s perspective on others’ understanding or misunderstanding of oneself; that is, an understanding of how one’s standpoint is perceived by others. At this stage, researchers can analyze the sociocultural implications of understanding and misunderstanding between subjects. \par

In this framework, self-identity serves as a foundational element. Researchers mediate dialogical communication across analytical levels, enabling an assessment of “the degree of convergence or divergence of perspective between Person A and B” (p. 23). \cite{Gillespie2010} note that while psychological traditions to intersubjectivity emphasize individual convergence, a sociological lens equally emphasizes community-level divergence, often manifesting as separation, division, or conflict. They argue that understanding a text’s embedded meanings requires a dialogue among different perspectives that reflects broader contexts. A text must be interpreted within a sociocultural context that transcends the immediate situations of its generation. Intersubjectivity, then, can be analyzed by comparing various perspectives at the first level, their relational understandings at the second, and their sociocultural contexts as “situation-transcending phenomena” (p. 33) at the third. \par

This dialogical approach aligns with \cite{Matusov1996}’s dialectical perspective, viewing intersubjectivity not as mere individual agreement but as a dynamic feature embedded in societal phenomena. Conceptualized this way, intersubjectivity becomes a process of coordinating varied perspectives, structured by dialectical relationships. Dialogical analysis, therefore, reveals both shared and divergent standpoints, mapping out possible relationships among them, fostering insights into diversifiable meanings of a text. \par

\section{Application of Dialectic Intersubjectivity in Content Analysis}
\label{sec:headings }
From \cite{Gillespie2010}'s approach, traditional content analysis centers on capturing first-level intersubjectivity through a consensus-oriented approach. Coders report their interpretations of a text based on researcher-provided guidelines. If significant disagreement emerges to a degree that threatens the measurement reliability (i.e., unacceptable intercoder agreement), researchers then systematically engage with second-level intersubjectivity. Here, although coders may not directly know how their direct perspectives converge or diverge from others, the researcher performs a kind of iterative meta-analysis to mediate understanding and misunderstanding among them \citep{riffe2023}. \par

Methodologically, traditional content analysis aims to enhance mutual understanding and reduce misunderstanding through refining coding instructions, coder retraining, and mediated discussions \citep{lombard2002, neuendorf2004, neuendorf2011, neuendorf2017}. Therefore, after resolving non-consensus, the third-level intersubjective communication of realizing both understanding and misunderstanding should not occur \citep{Gillespie2010}. Of course, such an agreement-seeking approach is indeed effective in capturing properties of texts accepted within consensus-oriented disciplines \citep{Krippendorff1999}, yet this imposes a paradox on analysts, requiring them to “have their own subjective interpretations”. \citep[ p. 269]{Potter1999} while excluding any attempt to “read between the lines”  \citep[ p. 18]{Berelson1952}. \par

Confronting this paradox, the dialectical approach diverges from traditional methods in the researcher’s communicative role at the second-level and in the presence of third-level communication. At the second-level, disagreement serves not merely as a source of misunderstanding, but as a tentative signal that various interpretations of the text might exist. Instead of treating unalignment as marginal errors failing to integrate into intersubjectivity, the researcher views these as potential generators of heterogeneous meanings. If misunderstandings are, in fact, identifiable as distinct interpretations—where they are reliably captured, respectively—then the researcher may proceed to assess intersubjectivity at the third-level. \par

At the third-level, the researcher identifies latent patterns among those perspectives. One homogeneous perspective sharing a common understanding may establish specific relations with another distinct perspective. These relations might be close vs. distant, similar vs. opposite, or reconciling or conflicting \citep{Lotman1988} and so on. By identifying those dialectic relations, the researcher coordinates individual perspectives while collectively contributing to the text’s heterogeneous meanings. This coordination is mediated through the researcher’s dialogical analysis, which posits that a text can radiate multifaceted meanings within a sociocultural context that transcends the immediate situation of its generation. In essence, the researcher seeks to understand the meaning(s) of the text, not as independent entities, but as interdependent relationships across various communities. \par

\begin{figure}[ht]
    \centering
    \includegraphics[width=0.8\linewidth]{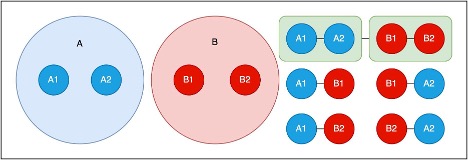}
    \caption{A Dialectic Model of Content Analysis}
    \label{fig:fig2}
    \vspace{1em} 
\end{figure}

Figure~\ref{fig:fig2} illustrates a model of dialectic content analysis, where intersubjectivity is conceptualized as a coordination of possible relationships among individual perspectives. In this model, the researcher starts by assuming that multiple distinct communities—in this case, Communities A and B—hold their own perspectives. To explore their variances, the researcher samples multiple coders from each community, with A1 and A2 for Community A, and B1 and B2 for Community B, to analyze the same text. These four coders receive uniform coding instructions and identical training, establishing a baseline of methodological consistency. \par

This setup creates six possible pairs of coder relationships, providing a means to examine how perspectives converge or diverge across both within-community and between-community comparisons. The researcher can assess Level 1 intersubjective relations by measuring the agreement or disagreement among these direct perspective pairs. Hypothetically, intercoder reliability should be higher for within-community pairs (A1-A2 and B1-B2), as coders from the same sociocultural group might share more similar interpretations of the text. This internal alignment demonstrates what \cite{Lotman1988} identifies as the first function of a text—fostering a unified reading among those with relatively similar backgrounds. However, this unifying function, as previously noted, is secondary to the second function of text analysis—enabling the creation of new meanings: While agreement among within-community pairs (A1-A2 and B1-B2) reflects shared understandings, it also suggests a tentative potential for divergent perspectives among between-community pairs (An-Bn). If intercoder reliability is consistently lower among these between-community pairs, the researcher might hypothesize that the text invites varying interpretations across different sociocultural contexts. \par

In traditional content analysis, coders from low-agreement pairs might be removed from the analysis, re-trained, or given more precise instructions in the codebook to achieve higher agreement \citep{riffe2023}. In contrast, the dialectic model posits that each coder is deeply embedded within a specific sociocultural context. Accordingly, disagreements, if they are systematically consistent, could serve as indicators of unintegrated standpoints. High intercoder reliability within A1-A2 and B1-B2, coupled with lower intercoder reliability within An-Bn, signals that Communities A and B may generate distinct valid interpretations of the same text through varying sociocultural lenses—a potential of Level 2 intersubjective relations. \par

Subsequently, the researcher should examine the relations between the within-community pairs to uncover latent patterns in how these communities interpret the text differentially. This relationship can be explored by assessing the degree of convergence or divergence in coding value distributions across each within-community coder. If the distributions show consistent but distinct patterns from each other, the researcher could identify sociocultural dynamics among coders’ views. This process allows to quantify the relationships among multiple perspectives on the same text, capturing what can be referred to as Level 3 intersubjective relations. \par

Together, the dialectic approach aims to quantitatively examine Level 3 intersubjectivity, a notion traditionally viewed as a limitation of content analysis or an impetus for subsequent qualitative approaches. Here, this third-level becomes a core objective, pushing content analysis beyond a consensus-seeking task and toward a dialectical joint activity aimed at uncovering sociocultural plurality within textual properties. \par

\section{Necessity of Dialectic Intersubjectivity in LLM-Assisted Content Analysis (LACA)}
\label{sec:headings }
We propose that embracing dialectical intersubjectivity is particularly relevant when using Large Language Models (LLMs) as instruments for content analysis. In LLM-based text analysis, the evaluation of coding performance typically relies on human-annotated datasets. Human annotations, regardless of their proliferation under terms such as “ground truth,” “golden standard,” or “benchmark” \citep{Heseltine2024, Wen2024}, often serve as an ultimate standard against which the utility of LLMs as tools for text analysis is judged. For instance, \cite{Gilardi2023} claim that ChatGPT classifications exhibit higher intercoder reliability with labels from human coders than those generated by two Mturk crowd workers, suggesting that “ChatGPT outperforms crowd workers for text-annotation tasks.” The underlying issue, however, is that this ground truth, derived from “two research assistants,” is neither unbiased nor error-free \citep{Platt2023}. \par

Ironically, this problem is compounded by the touted advantages of LLMs as a scientific tool: low cost, time efficiency, and scalability. If a single ground truth reflects only a specific, potentially biased perspective, then LACA’s scientific applications risk rapidly replicating these biases at scale (i.e., “reliably wrong”; \citealp{Song2020}, p. 555). Furthermore, if researchers aim for an artificial consensus, human bias and algorithmic bias may reinforce, rather than counterbalance, each other—leaving undetected errors unresolved (see \citealp{Giorgi2024}). \par

\cite{Messeri2024} encapsulate this potential risk in scientific AI applications as the illusion of understanding: when researchers assume that AI represents neutral or inclusive perspectives, its biases may go unnoticed, fostering a scientific monoculture that excludes alternative standpoints. They argue that addressing these epistemic risks requires embracing “both demographic diversity (and different attendant life experiences) and cognitive diversity (arising from different disciplinary training, skills, and problem-solving strategies)” (p. 54). Such diversity of standpoints enables multiple interpretations of AI analyses and minimizes the potential for AI applications to inadvertently erase human diversity (see also \citealp{Oreskes2019}). \par

In short, they emphasize that algorithmic bias is not an isolated issue but one shaped by the existing biases in society and research practices that accept manifested standpoints as sole truths. Likewise, if high intercoder agreement between LLM coding and human annotation reflects alignment with a particular standpoint, the analysis may inadvertently foster an illusion of understanding of the text. As \cite{Krippendorff1999} aptly noted, “all computer text analyses start with the identification of character strings, not meanings” (p. 54), and that “[e]ven perfectly dependable mechanical instruments [...] can be wrong—reliably” (p. 213). \par

In this light, we propose that integrating the concept of dialectical intersubjectivity into the LACA framework may provide a more valid delineation of research practices. By promoting the analysis not only of individual perspectives but also of meta-relations between them, the dialectical approach could provide LACA with workflows for inferring intricate meanings. \par

\section{Algorithmic Bias in LACA}
\label{sec:headings }
To this end, LACA must focus on two primary goals: (1) identifying perspectives an LLM tends to represent (e.g., \citealp{Santurkar2024}) and (2) assessing its ability to reflect other perspectives (e.g., \citealp{Argyle2023}). These priorities align with \cite{Bail2024}'s review of two major pathways for generative AI’s use in social sciences. \par

First, social scientists are developing frameworks for LLM-based textual analysis, based on the idea that these models could potentially mimic human intellectual processing \cite{Hancock2020, Mitchell2019}. This approach aims to harness LLMs in analyzing complex text (e.g., \citealp{Chew2023, Heseltine2024, Tornberg2023a, Ziems2024}). The second pathway involves using LLMs to support simulation-based research. Here, social scientists employ the agent-based modeling (ABM) paradigm, creating synthetic societies to observe interactions among simulated LLM personas \citep{Bail2024}. Studies in this area explore the potential of these models to predict real-world dynamics (e.g., \citealp{Park2023}). \par

Both approaches emphasize the potential for automated social science \citep{Demszky2023, Manning2024, stokelwalker2023}, while also recognizing algorithmic biases as a critical limitation. If algorithmic bias is viewed as an inherent cost of automated social science, then a key task becomes identifying which social groups’ standpoints an LLM may overrepresent and determining the extent to which this tendency can be adjusted to avoid misrepresenting other groups. Accordingly, for our purposes, it is crucial to examine how LLMs reflect specific perspectives and the extent to which they can simulate a diversity of standpoints across relevant social groups. \par

A prerequisite for these efforts is the conceptualization of algorithmic bias. \cite{Argyle2023} argue that LLMs exhibit multiple-interwoven biases rather than a single overarching bias, describing it as “a complex reflection of the many various patterns of association between ideas, attitudes, and contexts present among humans,” rather than “a singular, macro-level feature of the model.” Their analysis shows that when GPT-3 is prompted with specific demographic personas based on American National Election Study (ANES) data, its responses often mirror views typical of them, a phenomenon they term “algorithmic fidelity”—the degree to which LLMs mirror human ideas and attitudes across social contexts. They suggest that low algorithmic fidelity risks misrepresenting certain subgroups (pp. 337-338). \par

Similarly, \cite{Santurkar2024} find that conditioning LLMs on demographic profiles from Pew Research’s American Trends Panel (ATP) data makes responses reflect the actual opinions of these profiled groups. Their findings also show that default LLM responses tend to align more with liberal views, consistent with prior studies (e.g., \citealp{Hartmann2023}). Collectively, these studies envision integrating the two primary strands of LLM application in social scientific research, enabling us to observe how text-analyses might converge or diverge when LLMs simulate personas from different sociocultural groups (see \citealp{Giorgi2024}). \par

\section{Sentiment Analysis of Partisan Media with Persona-based LLMs}
\label{sec:headings }
In this exploratory study, we explore how partisan dispositions may shape interpretations of partisan texts and contexts; that is, we attempt to analyze not only how political content is encoded, but also how it is differentially decoded based on divergent partisan standpoints (see \citealp{Hall1980}), and ultimately, how these partisan standpoints are situated within the broader sociocultural context of political landscapes. To this end, we employ persona-based LACA, enabling a comparative exploration of how Democrat and Republican personas assess sentiment associated with Joe Biden and Donald Trump in transcripts from two prominent U.S. media outlets, Fox News and MSNBC, during the 2020 presidential campaign. \par

In line with this framework, we utilized six LLM configurations, leveraging OpenAI’s latest flagship model, gpt-4o-2024-08-06, at the time of analysis (the 4th quarter of 2024). Three models were further fine-tuned on the ANES 2020-2022 dataset, as OpenAI suggests that fine-tuning with relevant domain data can enhance performance on specialized tasks \citep{openai2024}. Thus, we utilized the latest ANES standard questionnaire capturing political orientations across U.S. demographics to improve the model’s ability to simulate partisan perspectives. Our models are configured as follows: (1) default zero-shot (no persona prompting) (DZ); (2) Democrat persona-prompted default (DD); (3) Republican persona-prompted default (DR); (4) fine-tuned zero-shot (FZ); (5) Democrat persona-prompted fine-tuned (FD); and (6) Republican persona-prompted fine-tuned (FR) models (see Supplemental Materials for the fine-tuning process). \par

\textit{RQ1: How do the six GPT-4o configurations evaluate sentiment positivity and negativity toward Joe Biden and Donald Trump within Fox News and MSNBC transcripts?}

\textit{RQ2: How does the degree of agreement vary across different model pairs?}

These questions reflect the first two levels of \cite{Gillespie2010}'s framework. By comparing sentiment assessments across models and examining intercoder reliability among pairs, we aim to identify the pairs that yield the most consistent interpretations of partisan views. Also, by determining which model most closely aligns with the outputs of the default zero-shot model, we may infer the personas alignment with GPT-4o’s baseline stance. \par

Finally, we expect that if intercoder reliability is indeed higher among models simulating the same partisan persona (i.e., within-community pairs: DD \& FD and DR \& FR), further investigation is warranted to explore differences in their analyses. Specifically, if persona-simulations exhibit partisan orientations, Democrat and Republican personas may diverge in their analyses of liberal versus conservative media. In other words, we anticipate that the congruence between the persona’s partisanship and the media source could produce distinct sentiment patterns toward each candidate. By analyzing how this Model-Text congruence (i.e., like-minded vs. cross-cutting) coordinates relationships among individual models’ analyses, the respective results can be examined meta-analytically within the sociocultural context of the polarized U.S. political landscape. This approach could flesh out the Level 3 intersubjectivity, realizing understanding/misunderstanding across partisan interpretations (see Figure ~\ref{fig:fig1}). \par

\textit{RQ3: How do sentiment analyses by partisan personas vary in their assessments of each candidate, depending on whether the source aligns with or opposes their partisanship?}

\section{Data and Methods}
\label{sec:others}
\subsection{Textual Data}
We utilized television transcripts from the (now deprecated) LexisNexis Web Services Kit. These are full transcripts for shows airing on Fox News and MSNBC. Because these networks do not use traditional evening news broadcasts in the way that ABC or CBS might, the corpus is all content from these networks that are regular and recurring shows. These data are described in more detail in (The author's citation is masked). \par

Fox News and MSNBC were selected as focal points for this analysis due to their well-documented contrast in political bias along the left-right spectrum \citep{AdFontes2024} and because the ANES 2020-2022 survey includes exclusive self-reported trust measures for these two networks. Since the data were used to fine-tune the default model of GPT-4o, analyzing content from these outlets aligns closely with the goal of enhancing algorithmic fidelity by reflecting political attitudes and opinions across U.S. demographics. \par

Fox News transcripts were available by broadcast unit, totaling 115,712 cases from 1998 to 2021, with an average word count of 2,798.7. During the campaign period from June to October 2020, 1,040 transcripts referenced both “Joe Biden” and “Donald Trump,” with an average word count of 7,294.4; MSNBC transcripts included 34,586 cases from 1999 to 2021, with an average word count of 7,408.8. Of these, 795 episodes from June to October 2020 referenced both candidates, averaging 8,103.4 words. Because LACA outcomes tend to vary with word count \citep{Heseltine2024}, transcripts with word counts between 7,500 and 8,000 were sampled, spanning the total average of 7,698.9 words across both sources. This yielded 242 MSNBC and 259 Fox News transcripts within the desired range. To equalize sample sizes, a random subsample of 242 cases was drawn from the Fox News data. \par

\subsection{Sentiment Analysis for Candidate Portrayals}
\subsubsection{Zero-shot Models}
Using the OpenAI API, we first assigned the role of assistant to the default zero-shot model (DZ) and prompted it to perform numeric sentiment coding for both Joe Biden and Donald Trump, with scores ranging from -2 for very negative, -1 for negative, 0 for neutral, 1 for positive, to 2 for very positive. The responses were requested in numeric form only. Given the relatively high word count per case, the maximum tokens per response were limited to 10, and each transcript was split into chunks of 2,000 tokens. To minimize randomness in the results, the temperature setting was fixed at 0 \citep{Santurkar2024,Tornberg2023a}. The same process was replicated using the fine-tuned zero-shot model (FZ). \par

\subsubsection{Persona-based Models}
Along with the same prompt and temperature setting, two partisan persona simulations were applied to both the default and fine-tuned models (DD, DR, FD, and FR). These personas were crafted by combining demographic attributes to reflect typical partisan profiles based on Pew Research’s recent reports \citep{Pew2024a,Pew2024b}. For Democrat persona, the prompt described: “a U.S. citizen who is a woman in her 20s, Black, with a college degree, Democrat, and middle income,” while for Republican persona, it described: “a U.S. citizen who is a man in his 50s, white, with a high school degree, Republican, and upper-middle income.” Each model was then prompted to analyze sentiment in the texts for each candidate based on these simulated personas (see Supplemental Materials for code and data). \par

\section{Results}
With respect to RQ1, which investigates how the six models evaluate sentiment toward Biden and Trump, our LACA surfaced the typical partisan biases of the two media outlets. Overall, all six models suggested that Fox News presented Biden in a negative light and Trump more favorably, whereas MSNBC exhibited the reverse pattern (Table 1). Yet, the Democrat persona models (DD and FD) outputted negative portrayals of Trump on Fox News as well. Conversely, the Republican persona fine-tuned model (FR) assessed Trump as being depicted even more favorably than Biden in MSNBC transcripts. \par

\begin{table}[ht]
\centering
\caption{Sentiment analysis for Biden and Trump in Fox News and MSNBC}
\label{tab:sentiment_analysis}
\begin{tabular}{@{}lcc|cc@{}}
\toprule
\textbf{Model} & \multicolumn{2}{c|}{\textbf{Fox News} $M$ ($SD$)} & \multicolumn{2}{c}{\textbf{MSNBC} $M$ ($SD$)} \\ 
               & \textbf{Biden}   & \textbf{Trump}  & \textbf{Biden}   & \textbf{Trump}  \\ \midrule
(1) DZ         & -0.34 (0.54) & 0.37 (0.64) & 0.16 (0.44) & -0.74 (0.74) \\
(2) DD         & -0.25 (0.52) & -0.03 (0.85) & 0.21 (0.52) & -0.61 (0.78) \\
(3) DR         & -0.59 (0.54) & 1.17 (0.49) & 0.03 (0.22) & -0.12 (0.78) \\
(4) FZ         & -0.30 (0.47) & 0.52 (0.58) & 0.21 (0.48) & -0.67 (0.74) \\
(5) FD         & -0.18 (0.41) & -0.08 (0.77) & 0.19 (0.47) & -0.96 (0.79) \\
(6) FR         & -0.42 (0.51) & 1.32 (0.38) & 0.06 (0.25) & 0.26 (0.70) \\ \bottomrule
\end{tabular}
\end{table}

Our analysis also explored the sentiment contrast between Biden and Trump across different models. We computed this by subtracting Trump’s sentiment score from Biden’s, yielding a range from $-4$ to $4$. Higher positive values signify a stronger positivity toward Biden over Trump, while lower negative values indicate a stronger negativity toward Biden compared to Trump. The one-way ANOVA results indicate that sentiment contrast differs significantly across models (for Fox News: $F(5, 1446) = 120.72$, $p < .001$, $\eta^2 = 0.294$, indicating a large effect; for MSNBC: $F(5, 1446) = 70.45$, $p < .001$, $\eta^2 = 0.197$, indicating a moderate effect) (see Table S1 for detailed post-hoc Tukey’s HSD analyses). \par

To address RQ2 of how agreement level differs among LLMs, intercoder reliability across fifteen pairs from six models was calculated using a random sample of 224 chunk responses (112 for each Fox News and MSNBC sample). This sampling approach aligns with recommended practices for large datasets, ensuring a representative assessment of coder consistency without requiring exhaustive computation of the entire dataset \citep{Krippendorff2018,lombard2002}. \par

As shown in Table 2, intercoder reliability metrics of Krippendorff’s $\alpha$ vary by how models are paired. The highest agreement was observed between default zero-shot and fine-tuned zero-shot models ($\alpha = .86$), followed by the pair of Democrat persona default and Democrat persona fine-tuned models ($\alpha = .85$). Notably, the pair of default zero-shot and Democrat persona default models showed acceptable intercoder reliability ($\alpha = .73$). This was even higher than the intercoder reliability from the pair of Republican persona default and Republican fine-tuned models, while this Republican pair showed a meritorious agreement ($\alpha = .68$) \citep{Hayes2007, Krippendorff2004}. Moreover, between-community pairs demonstrated consistent disagreement, with some pairs even showing negative intercoder reliability values. \par

\begin{table}[ht]
\centering
\caption{Intercoder reliability across model pairings}
\label{tab:intercoder_reliability}
\begin{tabular}{@{}lccccc@{}}
\toprule
\textbf{Model} & \textbf{DZ}    & \textbf{DD}                       & \textbf{DR}                       & \textbf{FZ}      & \textbf{FD}            \\ \midrule
\textbf{DZ}    &                &                                   &                                   &                  &                        \\
\textbf{DD}    & \textbf{0.73}  &                                   &                                   &                  &                        \\
\textbf{DR}    & 0.36  & \cellcolor{gray!30}0.07           &                                   &                  &                         \\
\textbf{FZ}    & \textbf{0.86}  & \textbf{0.66}                     & 0.44                     &                  &                          \\
\textbf{FD}    & \textbf{0.66}  & \cellcolor{blue!20}\textbf{0.85}  & \cellcolor{gray!30}-0.00          & 0.55    &                          \\ 
\textbf{FR}    & 0.16  & \cellcolor{gray!30}-0.10          & \cellcolor{red!20}\textbf{0.68}   & 0.26    & \cellcolor{gray!30}-0.18 \\ \bottomrule    
\end{tabular}

\begin{flushleft}
\textit{Note.} Cell entries are Krippendorff’s $\alpha$ for ordinal scales. Acceptable agreements are in bold. Blue-shaded for Democrat pair, Red-shaded for Republican pair, and Gray-shaded for cross-partisan pairs.
\end{flushleft}
\end{table}

Such differences in intercoder reliability between within-community and between-community pairs suggest that individual models may evaluate the same transcripts differently. To understand these simulated mutual misunderstandings within a sociocultural context of partisan polarization, we examined how persona-based LLMs vary in the convergence or divergence of sentiment contrast by media sources. This approach allowed us to scrutinize whether a persona’s evaluation of political candidates—pre-trained with data beyond the given materials—produces differences in sentiment contrast in analysis (RQ3). Specifically, to identify disparities in sentiment contrast distributions between the Democrat persona fine-tuned model (FD) and the Republican persona fine-tuned model (FR), we conducted Wasserstein Distance calculations and Kernel Density Estimation (KDE)\footnote{The Wasserstein distance quantifies the cost of transforming one probability distribution into another, providing a rigorous measure of their differences \citep{peyre2019,lombard2002}.}. \par

\begin{figure}
    \centering
    \includegraphics[width=0.8\linewidth]{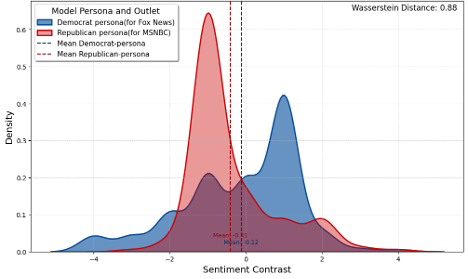}
    \caption{Distributions of Sentiment Contrast in Cross-cutting Transcript Analysis}
    \label{fig:fig3}
\end{figure}

First, for partisan incongruent analysis (Fox News analyzed by FD and MSNBC by FR), we observed significant overlap in sentiment contrast distributions (Figure ~\ref{fig:fig3}). In detail, the KDE showed that FD’s analysis of Fox News tended to be less negative toward Biden and less positive toward Trump, while FR’s analysis of MSNBC showed a tendency to be less negative toward Trump and less positive toward Biden. Consequently, the sentiment contrast distributions of these models converged rather than diverged, resulting in a Wasserstein distance of 0.88 on -4 to 4 scale. Given that the maximum possible distance on this scale is 8, this distance represents approximately 11\% of the maximum, indicating a small to moderate difference. \par

\begin{figure}
    \centering
    \includegraphics[width=0.8\linewidth]{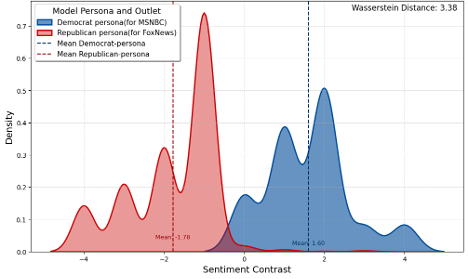}
    \caption{Distributions of Sentiment Contrast in Like-minded Transcript Analysis}
    \label{fig:fig4}
\end{figure}

In contrast, for partisan congruent analysis (MSNBC analyzed by FD and Fox News by FR), the sentiment contrast distributions formed by the two models exhibited substantial divergence (Figure ~\ref{fig:fig4}): KDE analysis revealed that FD’s analysis of MSNBC transcripts was more positive toward Biden and more negative toward Trump, while FR’s analysis of Fox News transcripts showed a stronger positive tendency toward Trump and a more negative tendency toward Biden. The Wasserstein distance of 3.38 represents roughly 42.25\% of the total possible range, indicating a considerable divergence. \par

\section{Discussion}
The utility of generative AI in social sciences is under scrutiny \citep{Bail2024}, with concerns around potential biases becoming pronounced as applications like LLM-Assisted Content Analysis (LACA) expand. These biases pose a challenge to the “ground truth” of human annotations, a persistent issue in traditional content analysis; that is, human biases could be amplified in LACA, further obscuring the complex meanings embedded in texts. To address these challenges, this study advocates for a conceptual shift from a consensus-oriented to a coordination-oriented intersubjectivity framework. Building on Matusov’s (1996) concept of dialectic intersubjectivity, we employ Gillespie and Cornish’s \citeyearpar{Gillespie2010} three-level model, which reframes intersubjectivity beyond mere consensus to include divergent perspectives. This model allows researchers to account for both agreements and productive disagreements, thereby capturing sociocultural nuances within a text. \par

Our findings underscore how dialectic intersubjectivity can reliably diversify LACA outputs, particularly when examining sentiment toward politically charged figures across partisan communities. First, sentiment analyses differ across LLM configurations, with within-community pairs (i.e., same partisan persona groupings) exhibiting higher intercoder reliability than between-community pairs. This suggests that partisan texts possess layers of meaning that are open to varied ideological interpretations, echoing Argyle et al.'s (2023) notion of “algorithmic fidelity”—how closely an LLM mirrors perspectives typical of specific social groups. Additionally, both the default and fine-tuned Democratic persona models consistently show higher agreement with the zero-shot GPT-4o than Republican personas, corroborating the liberal bias identified in prior research (e.g., Santurkar et al., 2023). This underscores the importance of accounting for algorithmic biases in LACA for political content. \par

Subsequent analysis of sentiment contrasts reveals notable variations between Democrat and Republican persona outputs depending on the media source. Democrat personas analyzing MSNBC and Republican personas analyzing Fox News demonstrate divergence in sentiment contrast toward Biden and Trump, while the same models analyzing cross-partisan content exhibit convergence. These findings suggest that selective exposure and processing do not function in isolation but rather form a dynamic system of meaning-making that is sociocultural context-dependent. Being reinforced by ideological alignment, such patterns manifest media-oriented political polarization \citep{prior2013} as a part of audience-oriented interpretive processes. Rather than merely receiving static encodings of “blue” or “red” media \citep{iyengar2009}, decoders may actively partake in dynamic coordination of various standpoints, navigating sociocultural interpretations of the text. \par

That said, the present dialectic approach does not reject traditional consensus-oriented approach’s strength in achieving reliable analysis but argues for a more relational outlook. By accommodating both agreement and disagreement, the dialectic approach opens new avenues: If content analysis forms, as Krippendorff (1999) analogizes, the “backbone” of communication research, its ribs must be further developed to support the discipline to stand upright. \par

Adopting a dialectical model does, however, present practical challenges. Recruiting a wide, demographically diverse group of coders to represent distinct sociocultural views is often unrealistic (see \citealp{lombard2002, neuendorf2011, neuendorf2017}. In this respect, LACA presents a compelling alternative by enabling the scalable simulation of various group profiles without the intensive resource requirements of conventional quantitative methods; that is, the LLM persona could facilitate the quantification of granular sociocultural contexts—an interpretative domain traditionally reserved for qualitative approach. Indeed, some may argue that the present work exemplifies how “LLMs challenge the conventional division between qualitative and quantitative methods in text analysis” \cite[p.5]{Tornberg2023a}. Yet, we contend that our approach not only preserves the qualitative domain but also fosters its ongoing evolution and deeper engagement. \par

For instance, recent research accumulates evidence rejecting the “filter bubble” and “echo chamber” hypotheses, indicating that digital media expose users to a broader range of opinions and sources than offline environments \citep{klinger2023}. Through the implications of our findings, users might encounter both polarizing like-minded content and depolarizing cross-cutting content simultaneously. If so, do these opposing effects neutralize one another, does one prevail, or do they interact to produce distinct outcomes? While \cite{tornberg2022} demonstrates that partisan-aligned information often overrides the depolarizing potential of cross-cutting content in digital media, the psychological and cognitive processes underlying this suppression remain unexplored. Amid these intersecting influences, when, under what conditions, and why do individuals ultimately revert to their partisan positions? This study fosters such novel qualitative inquiries of these dynamics. Moreover, integrating granular qualitative insights could refine agent-based model predictions of polarization dynamics (see Gao et al., 2023; Park et al., 2023). \par

In a similar vein, the present LACA raises a question about whether human ground truth can indeed be displaced by its simulacrum. In his early anticipations of AI, \cite{popper1978} underscores the function of human cognition in producing empirical knowledge, stating, “[h]uman understanding, and thus the human mind, seems to be quite indispensable” (p. 164). Although LLMs may simulate interpretive patterns resembling human cognitive work, they lack the experiential foundation required for empirical knowledge. Conscious AI remains an aspiration rather than a reality—an insight that proves relevant to the current findings (p. 165). \par

Future research might further test dialectical intersubjectivity by comparing outputs from LLM personas with those of human coders who provide equivalent profiles. Such studies could ascertain whether LLM-simulated intersubjectivity meaningfully translates to real-world phenomena, providing an empirical benchmark. This exploratory study lays the groundwork for a scalable, nuanced approach to incorporating diverse sociocultural contexts in LACA, yet continued research is essential to validate the simulated dialectics in real-world applications. \par

\newpage
\bibliographystyle{apalike}
\bibliography{references.bib} 

\newpage

\appendix 

\section*{Supplemental Material}

Our code and data are available at \href{https://github.com/casllmproject/dialectic_intersubjectivity}{https://github.com/casllmproject/dialectic\_intersubjectivity}.

\subsection*{A. Measurement scales for fine-tuning training data from the 2020-2022 ANES Social Media Study (American National Election Studies, 2023)} 

\subsubsection*{1. Demographics}

\begin{itemize}[label=$\bullet$]
    \item \textbf{Gender} (\textit{'What is your gender?'}):
    \begin{itemize}[label=$\circ$]
        \item 0: Unknown
        \item 1: Male
        \item 2: Female
    \end{itemize}

    \item \textbf{Age} (\textit{``What is your age?''}):
    \begin{itemize}[label=$\circ$]
        \item (numeric input)
    \end{itemize}

    \item \textbf{Race/Ethnicity} (\textit{``What is your race/ethnicity?''}):
    \begin{itemize}[label=$\circ$]
        \item 1: White, non-Hispanic
        \item 2: Black, non-Hispanic
        \item 3: Other, non-Hispanic (includes Asian, non-Hispanic)
        \item 4: Hispanic
    \end{itemize}

    \item \textbf{Education Level} (\textit{"What is your highest level of education?"}):
    \begin{itemize}[label=$\circ$]
        \item 1: Less than high school
        \item 2: High school graduate or equivalent
        \item 3: Vocational/tech school/some college/associates
        \item 4: Bachelor's degree
        \item 5: Postgraduate study/professional degree
    \end{itemize}
    \item \textbf{Income Level} (\textit{"What is your income level?"}):
    \begin{itemize}[label=$\circ$]
        \item 1: Less than \$5,000
        \item 2: \$5,000 to \$9,999
        \item 3: \$10,000 to \$14,999
        \item 4: \$15,000 to \$19,999
        \item 5: \$20,000 to \$24,999
        \item 6: \$25,000 to \$29,999
        \item 7: \$30,000 to \$34,999
        \item 8: \$35,000 to \$39,999
        \item 9: \$40,000 to \$49,999
        \item 10: \$50,000 to \$59,999
        \item 11: \$60,000 to \$74,999
        \item 12: \$75,000 to \$84,999
        \item 13: \$85,000 to \$99,999
        \item 14: \$100,000 to \$124,999
        \item 15: \$125,000 to \$149,999
        \item 16: \$150,000 to \$174,999
        \item 17: \$175,000 to \$199,999
        \item 18: \$200,000 or more
    \end{itemize}
\end{itemize}

\subsubsection*{2. Political Ideology}
\begin{itemize}[label=$\bullet$]
   \item \textbf{Self-Identified Political Ideology} (\textit{"When it comes to politics, would you describe yourself as liberal, conservative, or neither liberal nor conservative?"}):
   \begin{itemize}[label=$\circ$]
       \item -7: No answer
       \item -6: Unit non-response
       \item -1: Inapplicable (legitimate skip)
       \item 1: Very liberal
       \item 2: Somewhat liberal
       \item 3: Closer to liberal
       \item 4: Neither liberal nor conservative
       \item 5: Closer to conservative
       \item 6: Somewhat conservative
       \item 7: Very conservative
       \item 77: Don't know
       \item 98: Skipped on web
       \item 99: Refused
   \end{itemize}
   \item \textbf{Party-Specific Political Ideology}:
   \begin{itemize}[label=$\bullet$]
       \item \textit{"When it comes to politics, would you describe the Democratic Party as liberal, conservative, or neither liberal nor conservative?"}
       \item \textit{"When it comes to politics, would you describe the Republican Party as liberal, conservative, or neither liberal nor conservative?"}
   \end{itemize}
   \begin{itemize}[label=$\circ$]
       \item -7: No answer
       \item -6: Unit non-response
       \item -1: Inapplicable (legitimate skip)
       \item 1: Very liberal
       \item 2: Somewhat liberal
       \item 3: Closer to liberal
       \item 4: Neither liberal nor conservative
       \item 5: Closer to conservative
       \item 6: Somewhat conservative
       \item 7: Very conservative
       \item 77: Don't know
       \item 98: Skipped on web
       \item 99: Refused
   \end{itemize}
\end{itemize}

\subsubsection*{3. Trust in News Sources}
\begin{itemize}[label=$\bullet$]
   \item \textbf{Trust in Fox News} (\textit{"How much do you think political information from Fox News can be trusted?"}):
   \begin{itemize}[label=$\circ$]
       \item -7: No answer
       \item -6: Unit non-response
       \item -1: Inapplicable (legitimate skip)
       \item 1: Not at all
       \item 2: A little
       \item 3: A moderate amount
       \item 4: A lot
       \item 5: A great deal
       \item 77: Don't know
       \item 98: Skipped on web
       \item 99: Refused
   \end{itemize}
   \item \textbf{Trust in MSNBC} (\textit{"How much do you think political information from MSNBC can be trusted?"}):
   \begin{itemize}[label=$\circ$]
       \item -7: No answer
       \item -6: Unit non-response
       \item -1: Inapplicable (legitimate skip)
       \item 1: Not at all
       \item 2: A little
       \item 3: A moderate amount
       \item 4: A lot
       \item 5: A great deal
       \item 77: Don't know
       \item 98: Skipped on web
       \item 99: Refused
   \end{itemize}
\end{itemize}

\newpage

\subsection*{B. Fine-tuning Processes with ANES 2020-2022} 

\begin{figure}[ht]
    \centering
    \includegraphics[width=0.9\linewidth]{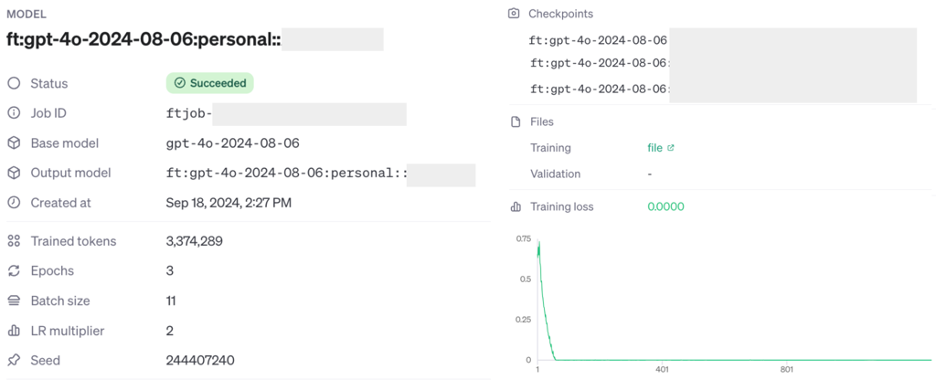}
    \captionsetup{labelformat=empty} 
    \caption{Figure S1. Screenshot of Fine-Tuning Job Results Display}
    \label{fig:S1}
    \vspace{1em} 
    \footnotesize 
    \textit{Note.} Information on the fine-tuned output model has been obscured to restrict public use.
\end{figure}

The training data for fine-tuning the gpt-4o-2024-08-06 base model consisted of the ANES 2020-2022 panel survey (n=5,750). The ten variables used for training included: gender, age, race/ethnicity, education level, income, self-identified political ideology, party-specific political ideology (Democratic and Republican parties), and trust in media (Fox News and MSNBC) (ANES, 2023). On September 18, 2024, the training data were uploaded and validated according to OpenAI’s fine-tuning protocol (OpenAI, n.d.). For the fine-tuning, the model was set with three epochs and a batch size of 11. Responses across the ten ANES variables totaled 3,374,289 tokens for the 5,750 respondent samples. The training process took approximately 2 hours and 58 minutes.

\newpage

\subsection*{C. Sentiment contrast analysis} 

\begin{table}[ht]
\centering
\caption{Tukey’s HSD analysis for mean difference of sentiment contrast across models}
\label{tab:S1}
\begin{tabular}{cc|ccc|ccc}
\hline
\multicolumn{2}{c|}{\textbf{Model}} & 
\multicolumn{3}{c|}{\textbf{Fox News Sentiment Contrast (Biden--Trump)}} & 
\multicolumn{3}{c}{\textbf{MSNBC Sentiment Contrast (Biden--Trump)}} \\
\hline
\textbf{I} & \textbf{J} & \textbf{Mdiff(I--J)} & \textbf{p-value} & \textbf{95\% CI} & \textbf{Mdiff(I--J)} & \textbf{p-value} & \textbf{95\% CI} \\
\hline
DZ & FD &  0.61 & <.001 & [0.35, 0.88] &  0.24 & 0.07  & [-0.01, 0.49] \\
DZ & DD &  0.49 & <.001 & [0.23, 0.76] & -0.09 & 0.90 & [-0.34, 0.16] \\
DZ & FZ & -0.11 &  0.86 & [-0.37, 0.16] & -0.03 & 0.999 & [-0.28, 0.22] \\
DZ & FR & -1.03 & <.001 & [-1.29, -0.76] & -1.11 & <.001 & [-1.36, -0.86] \\
DZ & DR & -1.04 & <.001 & [-1.31, -0.78] & -0.76 & <.001 & [-1.01, -0.51] \\
FD & DD &  0.12 &  0.79 & [-0.38, 0.14] & -0.33 & <.01  & [-0.58, -0.08] \\
FD & FZ & -0.72 & <.001 & [-0.98, -0.46] & -0.27 & <.05  & [-0.52, -0.02] \\
FD & FR & -1.64 & <.001 & [-1.90, -1.37] & -1.35 & <.001 & [-1.60, -1.10] \\
FD & DR & -1.66 & <.001 & [-1.92, -1.39] & -1.00 & <.001 & [-1.25, -0.75] \\
DD & FZ & -0.60 & <.001 & [-0.86, -0.34] &  0.06 & 0.982 & [-0.19, 0.31] \\
DD & FR & -1.52 & <.001 & [-1.78, -1.26] & -1.01 & <.001 & [-1.27, -0.76] \\
DD & DR & -1.54 & <.001 & [-1.80, -1.27] & -0.67 & <.001 & [-0.92, -0.42] \\
FZ & FR & -0.92 & <.001 & [-1.18, -0.66] & -1.08 & <.001 & [-1.33, -0.83] \\
FZ & DR & -0.94 & <.001 & [-1.20, -0.67] & -0.73 & <.001 & [-0.98, -0.48] \\
FR & DR & -0.02 & 1.000 & [-0.28, 0.25] &  0.35 & <.01  & [0.10, 0.60] \\
\hline
\end{tabular}
\end{table}

Post-hoc Tukey’s HSD analyses show that persona prompting significantly intensifies partisan bias for both outlets. However, this increase in bias does not vary significantly between default and fine-tuned models. Notably, the Fox News analysis showed no significant difference in sentiment contrast between the default and fine-tuned models. Similarly, for MSNBC, the default model’s sentiment contrast did not differ significantly from the fine-tuned zero-shot model or from either the default or fine-tuned Democrat persona models. In other words, the overall polarity of sentiment analysis for Biden and Trump remains consistent across these Default-to-Democrat model variations.

\end{document}